\documentclass{article}
\usepackage[preprint]{colm2026_conference}  

\usepackage[T1]{fontenc}
\usepackage{microtype}
\usepackage{hyperref}
\usepackage{url}
\usepackage{booktabs}
\usepackage{lineno}
\usepackage{graphicx}
\usepackage{subcaption}
\usepackage{tikz}
\usetikzlibrary{positioning, arrows.meta, shapes.geometric, calc, decorations.pathmorphing, shapes.callouts, backgrounds}

\definecolor{darkblue}{rgb}{0, 0, 0.5}
\hypersetup{colorlinks=true, citecolor=darkblue, linkcolor=darkblue, urlcolor=darkblue}

\newif\ifanon \anonfalse  

\title{Doing What They Say, Not What They Reason: Locating the
Faithfulness Gap in LLM Agents}

\author{Yufeng Wang \\
  \texttt{louiswang524@gmail.com}
}

\begin{document}

\maketitle

\begin{abstract}
Do LLM agents act on the reasoning they state? This question of process
fidelity is central to LLM-based social simulation, yet hard to measure where
no reference for correct behavior exists.
We study it in a controlled setting---a Texas Hold'em simulator with a
verifiable reference action for every decision---by splitting the faithfulness
gap into two steps: reasoning$\rightarrow$conclusion (does the stated decision
follow from the agent's own reasoning?) and conclusion$\rightarrow$action (does
the agent execute what it states?).
The two steps behave very differently.
Conclusion$\rightarrow$action is reliable: inconsistency is 0.7\% for Claude
Haiku 4.5 and 1.4\% for DeepSeek-Reasoner (a natively trained reasoning model) once
the conclusion is read from an explicit tag, whereas free-text conclusion
extraction reports 22--26\%---largely an artifact of how the conclusion is
parsed, not a property of the model.
Reasoning$\rightarrow$conclusion is where fidelity frays, but not through a
single dominant failure. In a step-level diagnostic the agent's errors split
roughly evenly between bad inputs, borderline cases, and \emph{rule
misapplication}---deriving a conclusion that contradicts the agent's own
restated rule from inputs it estimated correctly (about a third of interpretable
errors for Haiku). This composition is model-dependent: rule misapplication
accounts for a third of Haiku's interpretable errors but only 8\% of
DeepSeek's. The one robust signal is directional: when an agent does misapply
its own stated rule, it almost always (99.5\% for Haiku) errs in the
risk-averse direction. The override is partly hedging behavior, not a capability
limit: instructing the agent to apply the rule mechanically halves the
misapplication rate (13.9\% to 6.8\% of decisions) and raises adherence by eight
points. Because these are internal-consistency checks, they do not depend on the
reference policy being correct.
Process-fidelity evaluation should therefore elicit machine-checkable
conclusions and probe for directional biases rather than assume a single
upstream failure mode, lest it conflate measurement noise with model behavior.
We release code and a live multi-agent demo.
\end{abstract}

\section{Introduction}

A fundamental concern in LLM-based social simulation is whether agents
genuinely act on the reasoning they appear to exhibit, or produce
surface-level rationalizations that do not govern behavior
\citep{park2023generative,zhou2024sotopia}.
This concern is sharpened by a growing critique that the field deploys
simulators faster than it validates them, and that reported results depend
heavily on the elicitation protocol and measurement instrument used
\citep{puelmatouzel2026validation}.
The question is difficult to resolve in open social settings because ground
truth is absent: no objective criterion exists for the correct action in a
negotiation or a cultural exchange.

Competitive games with a computable reference policy offer a different regime.
In Texas Hold'em poker, a reasonable reference action at each decision point
can be derived from hand equity and pot odds---a fixed reference that is
unavailable in most social simulation contexts.
This property renders poker a controlled calibration environment for
\emph{process fidelity}: when an LLM agent claims to reason about pot odds and
hand strength, the executed action, the stated conclusion, and the
intermediate reasoning can each be inspected against one another and against
the reference.
Open-ended social simulations---marketplaces, negotiations, multi-agent
societies---lack this verifiability, yet the agents that populate them are the
same models, prompted in the same way. A failure mode that is measurable in
poker is therefore a candidate mechanism for the harder-to-measure breakdowns
of persona consistency and emergent behavior reported in richer settings such
as SOTOPIA \citep{zhou2024sotopia}.

Using this setup, we split the ``faithfulness gap'' into two steps that turn
out to behave very differently.
\emph{Conclusion$\rightarrow$action} asks whether the agent executes the
decision it states; \emph{reasoning$\rightarrow$conclusion} asks whether that
stated decision follows validly from the agent's own intermediate reasoning.
Prior work has reported that CoT explanations are frequently unfaithful
\citep{turpin2023language} and that LLMs exhibit a ``knowing-doing gap'' in
game-theoretic settings \citep{lin2026far}; our results refine this picture by
locating the gap almost entirely in the second step.
Across three LLM families (Claude Haiku 4.5, Gemini 2.5 Flash-Lite, and
DeepSeek-Reasoner---the last a natively trained reasoning model) and four prompt
strategies, we address two research questions:

\begin{itemize}
  \item \textbf{RQ1}: How large is the conclusion$\rightarrow$action
    inconsistency rate under game-theoretic pressure, and how sensitive is the
    measured rate to how the conclusion is elicited and parsed?
  \item \textbf{RQ2}: When the reasoning is made explicit, do agents derive
    conclusions that follow from their own stated inputs and rules; and does
    injecting the decision rule into the prompt improve adherence to it?
\end{itemize}

\paragraph{Contributions.}
\begin{itemize}
  \item \textbf{A measurement caution for CoT faithfulness.} The measured
    stated-vs.-actual inconsistency rate is an order of magnitude lower under
    explicit-conclusion elicitation (0.7\% for Haiku, 1.4\% for DeepSeek) than
    under free-text phrase extraction (22--26\%), indicating that a substantial
    part of an apparent CoT ``unfaithfulness'' signal can reflect
    conclusion-extraction noise rather than model behavior.
  \item \textbf{A decomposition that locates the gap.} Separating
    reasoning$\rightarrow$conclusion from conclusion$\rightarrow$action shows the
    two steps behave very differently: agents execute their stated decisions
    almost perfectly, whereas the derivation of those decisions from stated
    reasoning is where fidelity frays.
  \item \textbf{A characterization of the upstream failure---its directionality,
    and a mitigation.} In a step-level diagnostic that elicits the agent's own
    inputs and rule, no single cause dominates: for Claude Haiku 4.5 the errors
    split roughly evenly across input errors, rule \emph{misapplication}, and
    borderline cases (each about a third). The robust signal is directional---
    when the agent does misapply its own restated rule, it almost always
    (99.5\%) errs risk-aversely. The override is partly hedging behavior:
    instructing the agent to apply the rule mechanically halves the
    misapplication rate and lifts adherence by eight points. The composition is
    also model-dependent---rule misapplication is a third of Haiku's
    interpretable errors but only 8\% of DeepSeek-Reasoner's---so the upstream gap
    is not a single universal failure mode. Because these are
    \emph{internal}-consistency checks (stated rule vs.\ stated decision), they
    are robust to the choice of reference policy.
  \item \textbf{A portable process-fidelity instrument.} We package poker as a
    verifiable, reproducible testbed with a step-level fidelity probe (open
    harness and live multi-agent demo) that can be ported to richer social
    simulations to detect persona drift and surface-level rationalization
    before they propagate.
\end{itemize}

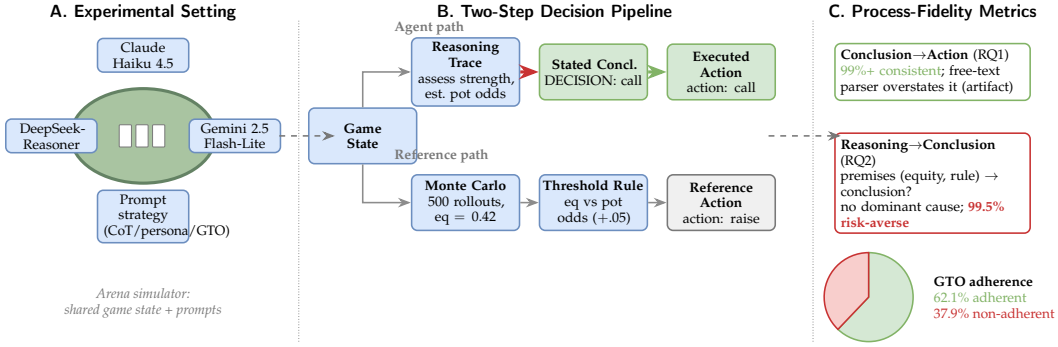
\begin{figure*}[t]
  \centering
  \definecolor{pastelblue}{RGB}{218, 232, 252}
  \definecolor{tikzblue}{RGB}{108, 142, 191}
  \definecolor{pastelgreen}{RGB}{213, 232, 212}
  \definecolor{darkgreen}{RGB}{130, 179, 102}
  \definecolor{pastelgray}{RGB}{240, 240, 240}
  \definecolor{darkgray}{RGB}{120, 120, 120}
  \definecolor{pastelred}{RGB}{248, 206, 204}
  \definecolor{darkred}{RGB}{200, 50, 50}
  \definecolor{tablegreen}{RGB}{169, 196, 161}
  \definecolor{tableborder}{RGB}{100, 130, 100}
  \resizebox{\textwidth}{!}{%
  \begin{tikzpicture}[
      >=Stealth, font=\sffamily,
      hdr/.style={font=\sffamily\bfseries\small, anchor=north},
      seat/.style={rectangle, rounded corners=3pt, draw=tikzblue,
          fill=pastelblue, text width=1.5cm, align=center, font=\scriptsize,
          minimum height=0.6cm, thick, inner sep=2pt},
      pbox/.style={rectangle, rounded corners=3pt, text width=1.8cm,
          align=center, font=\scriptsize, minimum height=1.0cm, thick,
          draw=tikzblue, fill=pastelblue, inner sep=2pt},
      gpbox/.style={pbox, draw=darkgreen, fill=pastelgreen},
      ypbox/.style={pbox, draw=darkgray, fill=pastelgray},
      cbox/.style={rectangle, rounded corners=3pt, text width=3.3cm,
          align=left, font=\scriptsize, minimum height=1.0cm, thick,
          inner sep=3pt, fill=white},
      arrow/.style={->, thick, draw=darkgray!85},
      greenarrow/.style={->, ultra thick, draw=darkgreen},
      redarrow/.style={->, ultra thick, draw=darkred},
      flow/.style={->, thick, dashed, draw=darkgray}
  ]
  \draw[densely dotted, draw=darkgray] (5.3,-2.3) -- (5.3,3.5);
  \draw[densely dotted, draw=darkgray] (14.55,-2.3) -- (14.55,3.5);
  \node[hdr] at (2.5,3.9) {A.\ Experimental Setting};
  \node[hdr] at (9.9,3.9) {B.\ Two-Step Decision Pipeline};
  \node[hdr] at (16.7,3.9) {C.\ Process-Fidelity Metrics};

  \node[ellipse, draw=tableborder, fill=tablegreen, minimum width=2.6cm,
        minimum height=1.7cm, ultra thick] (tbl) at (2.5,1.4) {};
  \foreach \x in {2.07,2.37,2.67} {
    \draw[fill=white, draw=darkgray, rounded corners=0.5pt]
      (\x,1.2) rectangle ++(0.22,0.4);
  }
  \node[seat] at (2.5,2.85)  {Claude Haiku 4.5};
  \node[seat] at (2.5,-0.05) {Prompt strategy\\\,(CoT/persona/GTO)};
  \node[seat] at (0.85,1.4)  {DeepSeek-Reasoner};
  \node[seat] at (4.15,1.4)  {Gemini 2.5 Flash-Lite};
  \node[font=\scriptsize\itshape, text=darkgray, align=center] at (2.5,-1.6)
    {Arena simulator:\\shared game state + prompts};

  \node[pbox] (gs) at (6.45,1.4) {\textbf{Game}\\\textbf{State}};
  \node[pbox]  (rt) at (8.3,2.55) {\textbf{Reasoning Trace}\\assess strength, est.\ pot odds};
  \node[gpbox] (sc) at (10.6,2.55) {\textbf{Stated Concl.}\\DECISION: call};
  \node[gpbox] (ea) at (12.9,2.55) {\textbf{Executed Action}\\action: call};
  \node[pbox]  (mc) at (8.3,0.2)  {\textbf{Monte Carlo}\\500 rollouts, eq $=0.42$};
  \node[pbox]  (tr) at (10.6,0.2) {\textbf{Threshold Rule}\\eq vs pot odds ($+.05$)};
  \node[ypbox] (ra) at (12.9,0.2) {\textbf{Reference Action}\\action: raise};
  \node[font=\scriptsize\bfseries, text=darkgray, anchor=west] at (6.9,3.35) {Agent path};
  \node[font=\scriptsize\bfseries, text=darkgray, anchor=west] at (6.9,1.05) {Reference path};
  \draw[arrow] (gs.north) |- (rt.west);
  \draw[arrow] (gs.south) |- (mc.west);
  \draw[redarrow]   (rt) -- (sc);
  \draw[greenarrow] (sc) -- (ea);
  \draw[arrow] (mc) -- (tr);
  \draw[arrow] (tr) -- (ra);

  \node[cbox, draw=darkgreen] (c1) at (16.7,2.55)
    {\textbf{Conclusion$\rightarrow$Action} (RQ1)\\
     {\color{darkgreen}99\%+ consistent}; free-text parser overstates it (artifact)};
  \node[cbox, draw=darkred] (c2) at (16.7,0.55)
    {\textbf{Reasoning$\rightarrow$Conclusion} (RQ2)\\
     premises (equity, rule) $\rightarrow$ conclusion?\\
     no dominant cause; {\color{darkred}\textbf{99.5\% risk-averse}}};
  \def\cx{15.55}\def\cy{-1.5}\def\rr{0.8}
  \fill[pastelgreen, draw=darkgreen, thick] (\cx,\cy) -- ++(90:\rr)
        arc[start angle=90, end angle=-133.6, radius=\rr] -- cycle;
  \fill[pastelred, draw=darkred, thick] (\cx,\cy) -- ++(-133.6:\rr)
        arc[start angle=-133.6, end angle=-270, radius=\rr] -- cycle;
  \node[font=\scriptsize, align=left, anchor=west] at (16.6,-1.5)
    {\textbf{GTO adherence}\\{\color{darkgreen}62.1\% adherent}\\{\color{darkred}37.9\% non-adherent}};

  \draw[flow] (4.95,1.4) -- (5.95,1.4);
  \draw[flow] (13.75,1.4) -- (15.05,1.4);
  \end{tikzpicture}%
  }
  \caption{Overview of the experimental framework.
    \textbf{(A)} Three LLM families---Claude Haiku 4.5, Gemini 2.5 Flash-Lite,
    and DeepSeek-Reasoner---plus a prompt-strategy seat play a shared Texas
    Hold'em table managed by the arena simulator.
    \textbf{(B)} Each decision is decomposed into two paths: an \emph{agent
    path} (reasoning trace $\rightarrow$ stated conclusion $\rightarrow$
    executed action) and a \emph{reference path} (Monte Carlo equity
    $\rightarrow$ threshold rule $\rightarrow$ reference action).
    \textbf{(C)} The two measured steps behave very differently:
    conclusion$\rightarrow$action consistency is near-perfect (RQ1; the
    free-text parser's apparent gap is a measurement artifact), whereas
    reasoning$\rightarrow$conclusion is where fidelity frays---with no single
    dominant error cause, but rule misapplications running 99.5\% in the
    risk-averse direction (Haiku), and overall GTO adherence at 62.1\%.}
  \label{fig:overview}
\end{figure*}

The answers bear directly on a central concern for LLM-based social
simulation: distinguishing genuine strategic behavior from model artifacts.
Apparent stated-vs.-actual unfaithfulness is, in this setting, an artifact of
how the conclusion is measured, while a less commonly measured failure---
invalid reasoning that the agent nonetheless executes faithfully---is both real
and substantial.

\section{Related Work}

\paragraph{Chain-of-thought faithfulness.}
\citet{wei2022chain} demonstrated that chain-of-thought prompting
substantially improves LLM performance on multi-step reasoning tasks.
\citet{turpin2023language} showed that CoT explanations can be unfaithful---
models influenced by spurious biasing features fail to mention those
influences while still acting on them---a finding established through careful
controlled manipulation rather than free-text conclusion extraction.
\citet{lanham2023measuring} measured faithfulness directly through
interventions on the reasoning trace (truncation, paraphrasing, added
mistakes), establishing that the \emph{measurement instrument} shapes the
conclusion one draws about faithfulness.
Our results complement this line of work with a methodological observation
specific to our setting: when the stated conclusion is \emph{inferred} from
free-text reasoning rather than controlled directly, the measured
stated-vs.-actual inconsistency rate is highly sensitive to the extraction
method, varying by an order of magnitude between one plausible naive parser (a
last-match phrase heuristic) and an explicit decision tag. The claim is about
this naive extraction method, not about published faithfulness rates, and it
concerns how the gap is operationalized, not whether CoT can be unfaithful.

\paragraph{LLM multi-agent social simulation.}
\citet{park2023generative} showed that LLM agents equipped with memory,
reflection, and planning can produce emergent social behaviors in sandbox
environments.
SOTOPIA \citep{zhou2024sotopia} introduced an interactive benchmark for
social intelligence, finding that GPT-4 achieves significantly lower
goal-completion rates than humans on challenging social scenarios.
\citet{puelmatouzel2026validation} survey this fast-growing area and argue that
it must pivot from expansion to consolidation around reproducible, validated
evaluation---calling for \emph{operational validity}, the property that a
simulator reproduces rather than merely resembles the target phenomenon.
These works largely evaluate \emph{outcome} quality, or argue for validation in
the abstract.
We complement them with a concrete, verifiable \emph{process}-fidelity
measurement at the level of individual reasoning steps.

\paragraph{Strategic reasoning and the knowing-doing gap.}
GameBench \citep{costarelli2024gamebench} evaluated LLMs across nine strategic
game environments, finding that none matched human performance.
\citet{lin2026far} examined LLMs in Texas Hold'em specifically, identifying a
``knowing-doing gap'' between reasoning traces and decisions, and addressed it
via external solver tools.
\citet{xie2026m3bench} introduced M3-BENCH and observed an
``overthink-undercommunicate'' pattern.
Our work localizes the knowing-doing gap: it is not that agents fail to
execute their stated decisions, but that they fail to derive correct
conclusions from reasoning they themselves produce.

\section{System and Methodology}

\subsection{The Agent Harness}

We construct a Texas Hold'em simulator in which every seat is occupied by an
LLM agent.
The harness comprises three components.
The \textbf{arena} (\texttt{engine/game.py}) manages shared game state---deck,
community cards, pot, stacks, and betting order---advancing after each agent
action.
The \textbf{agent interface} (\texttt{ai/llm\_player.py}) exposes a single
\texttt{decide(game) $\rightarrow$ action} method.
The \textbf{observer layer} (\texttt{experiment/}) logs every decision
alongside a reference action from an equity-based threshold heuristic,
enabling process-level evaluation independent of chip outcomes.
Stacks are reset to 10{,}000 chips (1{,}000 big blinds) at the start of each
hand, rendering each hand an independent observation and ensuring equal
participation across all four strategy seats.

\subsection{Prompt Strategies}

Each seat is assigned one of four system prompt strategies, held constant
throughout the experiment. Every strategy receives the same per-decision user
prompt describing the game state (Table~\ref{tab:prompts}, bottom) and differs
only in its system prompt. The strategies are designed to vary one factor: how
much, and what kind of, decision guidance is supplied externally.

\begin{itemize}
  \item \textbf{Baseline}: a minimal expert framing with no reasoning
    requested. The model is told only to ``accumulate chips by making +EV
    decisions'' and to ``respond with valid JSON only.'' This isolates the
    model's default behavior absent any elicited reasoning.
  \item \textbf{CoT}: identical framing, but the model is instructed to
    ``think step by step: assess your hand strength, estimate pot odds,
    consider opponents,'' then to emit an explicit
    \texttt{DECISION: <fold|check|call|raise>} line before the JSON. The
    explicit decision line is what makes the conclusion machine-checkable
    (Section~4.1).
  \item \textbf{Persona}: a behavioral style is imposed rather than a
    reasoning procedure: ``You are a tight-aggressive (TAG) player\ldots play
    only premium hands aggressively, fold marginal hands, and apply pressure
    from position.'' No reasoning is requested.
  \item \textbf{GTO-constrained}: the decision rule itself is injected into the
    prompt: ``estimate equity vs pot odds\ldots fold if equity $<$ pot odds;
    call or raise if equity $>$ pot odds.'' This tests whether handing the
    model the rule improves adherence to it.
\end{itemize}

For the reasoning$\rightarrow$conclusion analysis we add a diagnostic strategy,
\textbf{GTO-verbose}, which instructs the agent to emit four labeled lines
before its JSON---\texttt{EQUITY}, \texttt{POT\_ODDS}, \texttt{RULE}, and
\texttt{DECISION}---so that the agent's own estimated inputs, its restatement
of the rule, and its stated decision can each be checked against the others and
against the Monte Carlo reference. The verbatim system prompts and an example
input are listed in Appendix~\ref{app:prompts}.

\subsection{Metrics}

\textbf{GTO adherence} is computed per decision by a Monte Carlo equity
calculator (500 simulations) combined with an equity-based threshold
heuristic: fold if equity $<$ pot\_odds; call if $|$equity $-$ pot\_odds$|$
$<$ 0.05; raise otherwise.
We employ this simplified heuristic as a reproducible, fixed reference point
rather than an exact game-theoretically optimal policy; exact equilibrium
computation in multi-player Hold'em is intractable at this scale.
It is applied identically to all agents and strategies.

\textbf{Conclusion$\rightarrow$action consistency} applies to the CoT and
GTO-verbose strategies, which emit an explicit \texttt{DECISION} tag.
A decision is flagged \emph{inconsistent} when the tagged conclusion
contradicts the JSON action.
To quantify the sensitivity of this measurement, we compare the explicit-tag
parser against a baseline regular-expression parser that infers the conclusion
from free-text phrases such as ``should raise'' or ``best to call''.

\textbf{Reasoning$\rightarrow$conclusion validity} applies to the GTO-verbose
strategy.
For each decision we check whether the stated \texttt{DECISION} follows from
the agent's own stated \texttt{EQUITY} and \texttt{POT\_ODDS} under the rule it
restated.

Figure~\ref{fig:overview}(B) summarizes how a single decision flows through the
two paths and where each measurement is taken. Note that the Monte Carlo equity
serves a dual role: it determines the reference action (for GTO adherence) and
provides the independent yardstick against which the agent's \emph{stated}
equity is checked (for reasoning$\rightarrow$conclusion validity), which is what
lets us separate input errors from rule misapplication.

\subsection{Experimental Setup}

Three independent 200-hand experiments are conducted with
\texttt{claude-haiku-4-5-20251001}, \texttt{gemini-2.5-flash-lite}, and
\texttt{deepseek-reasoner}, each at the model-default temperature.
The reasoning$\rightarrow$conclusion diagnostic (Section~4.2) uses a separate
GTO-verbose run in which all four seats run the verbose strategy on one model:
150 hands (2{,}032 decisions) on \texttt{claude-haiku-4-5-20251001} and 200 hands
(2{,}036 decisions) on \texttt{deepseek-reasoner}.
Because \texttt{deepseek-reasoner} bills its chain of thought as completion
tokens, we set its completion budget to 8{,}192 so the reasoning does not crowd
out the labeled answer.
Starting stack: 10{,}000 chips; big blind: 10 chips; Monte Carlo simulations:
500 per decision; stacks are reset to the starting amount at the start of each
hand so every hand is an independent observation.
Each main run places one seat per strategy at a single table; all four seats use
the same underlying model.
Code and decision logs are available \ifanon
\else
at \url{https://github.com/louiswang524/texas_poker}.
\fi

\section{Results}

\begin{figure*}[t]
  \centering
  \begin{subfigure}[t]{0.40\textwidth}
    \centering
    \includegraphics[width=\textwidth]{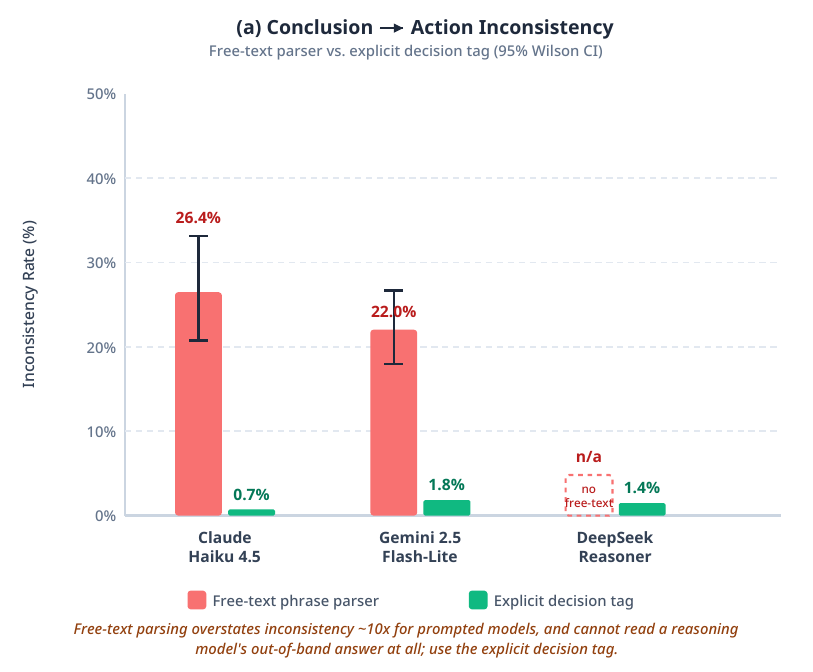}
  \end{subfigure}
  \hfill
  \begin{subfigure}[t]{0.58\textwidth}
    \centering
    \includegraphics[width=\textwidth]{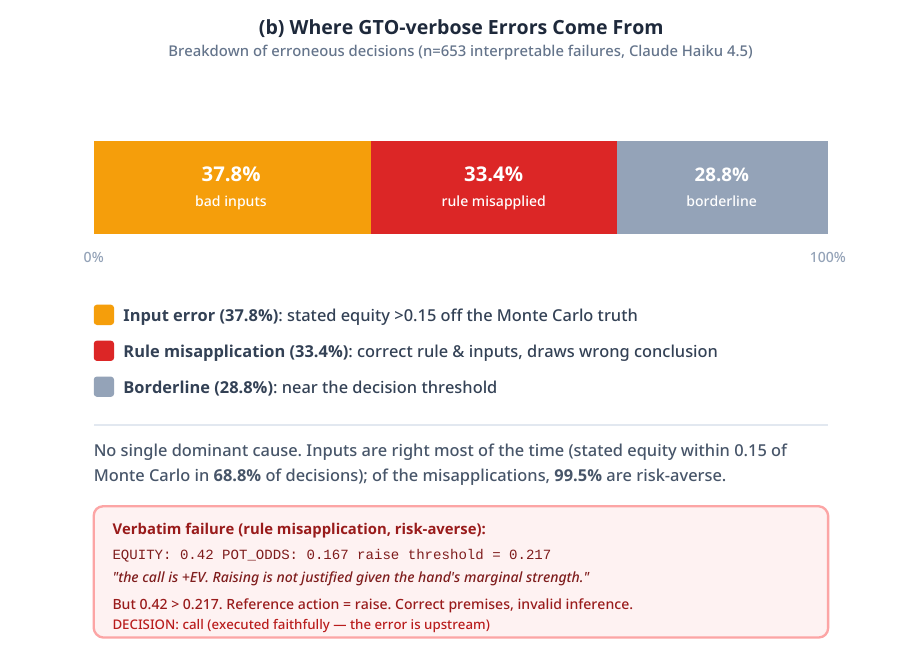}
  \end{subfigure}
  \caption{%
    \textbf{(a)} Conclusion$\rightarrow$action inconsistency under two parsers.
    A free-text phrase parser reports 22--26\%; an explicit decision tag
    reduces the same quantity to under 2\%, revealing the former as a
    measurement artifact.
    \textbf{(b)} Where erroneous GTO-verbose decisions come from (Claude Haiku
    4.5): they split roughly evenly across bad inputs, misapplication of a
    correctly stated rule, and borderline cases, with no single dominant cause.
  }
  \label{fig:results}
\end{figure*}

\subsection{Conclusion$\rightarrow$Action Inconsistency Is a Measurement
Artifact (RQ1)}

Table~\ref{tab:inconsistency} reports the inconsistency between the stated
conclusion and the executed action under two parsers
(Figure~\ref{fig:results}a).
With a free-text phrase parser, the rate is 26.4\% (Haiku) and 22.0\%
(Flash-Lite), and a large fraction of decisions cannot be parsed at all.
With an explicit \texttt{DECISION} tag, the inconsistency collapses to 0.7\% and
1.8\% respectively, recovered in 95--100\% of these two prompted models'
responses.
DeepSeek-Reasoner is a separate case. Because it returns its chain of thought in
a dedicated reasoning field rather than in the answer text, its answer carries an
extractable free-text conclusion in under 1\% of responses---there is essentially
nothing for the phrase parser to read---so the free-text inconsistency is not
meaningful and only the explicit tag is usable. Read from that tag, its
inconsistency is 1.4\% over the 62\% of responses that emit the tag before the
turn ends (Section~4.2).

\begin{table}[h]
\centering
\caption{Conclusion$\rightarrow$action inconsistency by parser, with 95\%
Wilson confidence intervals. The free-text parser overstates inconsistency by
roughly an order of magnitude. $^{\dagger}$DeepSeek-Reasoner returns its chain of
thought in a separate field, so an extractable free-text conclusion appears in
under 1\% of its answers (parse rate 0.7\%, $n=3$); its free-text inconsistency
is therefore not meaningful, and only the explicit tag is usable.}
\label{tab:inconsistency}
\begin{tabular}{lcccc}
\toprule
& \multicolumn{2}{c}{Free-text phrase parser} & \multicolumn{2}{c}{Explicit decision tag} \\
\cmidrule{2-3} \cmidrule{4-5}
Model & Inconsistency & Parse rate & Inconsistency & Parse rate \\
\midrule
Claude Haiku 4.5      & 26.4\% {[}20.7, 33.1{]} & 51\% & 0.7\% {[}0.4, 1.2{]} & 95\% \\
Gemini 2.5 Flash-Lite & 22.0\% {[}18.0, 26.6{]} & 76\% & 1.8\% {[}0.9, 3.5{]} & 100\% \\
DeepSeek-Reasoner       & n/a$^{\dagger}$ & 0.7\% & 1.4\% {[}0.9, 2.2{]} & 62\% \\
\bottomrule
\end{tabular}
\end{table}

The agents almost always execute the decision they state.
The apparent unfaithfulness reported by the phrase parser arises because
free-text reasoning mentions several candidate actions in passing
(``I could call here, but raising is stronger''), and a last-match heuristic
frequently extracts the wrong one.
This answers RQ1: the conclusion$\rightarrow$action gap is small, and the
larger rates reported by free-text phrase parsing reflect conclusion
extraction.

\subsection{Verbose Errors Are Diverse in Cause but Uniform in Direction (RQ2)}

The GTO-verbose diagnostic exposes the agent's own estimated inputs and restated
rule. We ran it on Claude Haiku 4.5 over 150 hands; of 2{,}032 decisions,
1{,}928 (94.9\%) produced a fully parsed labeled response, and all figures below
are over those.\footnote{The remaining 5.1\% were truncated or non-compliant and
are excluded rather than scored on a fallback action. A second, smaller run on
DeepSeek-Reasoner is reported at the end of this subsection.}
Overall adherence to the Monte Carlo reference action is 62.1\% (95\% Wilson CI
[59.9, 64.2]), leaving 731 erroneous decisions, of which 653 have all four
labeled lines parsed (the interpretable subset).
Crucially, the breakdown that follows is an \emph{internal}-consistency
check---the agent's stated \texttt{DECISION} against its own stated
\texttt{EQUITY}, \texttt{POT\_ODDS}, and restated \texttt{RULE}---so it does not
depend on the reference policy being correct, and is therefore robust to the
limitations of our adherence heuristic (Section~3.3).

No single cause dominates the errors. They split into three comparable groups
(Figure~\ref{fig:results}b): \emph{input errors}, where the stated equity is
more than 0.15 from the Monte Carlo value (37.8\%, 95\% CI [34.2, 41.6]);
\emph{rule-misapplications}, where the agent states plausible inputs and the
correct rule but announces a decision that does not follow from them (33.4\%,
[29.9, 37.1]); and borderline near-threshold cases (28.8\%, [25.4, 32.4]). The
inputs are right most of the time---the stated equity is within 0.15 of the
Monte Carlo value in 68.8\% (95\% CI [66.7, 70.9]) of decisions that state
one---so misapplied reasoning is a genuine error mode, but it is one of three
comparable sources rather than the dominant one.

A representative rule-misapplication, reproduced verbatim from the logs:
\begin{quote}\small
\texttt{EQUITY: 0.42} \quad \texttt{POT\_ODDS: 0.167} \\
\texttt{RULE: fold if EQUITY < POT\_ODDS; raise if EQUITY > POT\_ODDS + 0.05;
otherwise call} \\
\texttt{DECISION: call} \\
``With 42\% equity and pot odds requiring only 16.7\% equity, the call is +EV.
Raising is not justified given the hand's marginal strength against multiple
opponents.''
\end{quote}
The rule the agent just restated implies \texttt{raise} ($0.42 > 0.167 + 0.05$),
and the reference action is \texttt{raise}; the agent instead overrides it with a
qualitative judgment of ``marginal strength.'' The failure is neither an input
error nor an execution error---it is an invalid inference from the agent's own
correctly stated premises.

The sharper finding is the \emph{direction} of these misapplications. If the
agent could not perform the numerical comparison, errors would be roughly
symmetric; instead they are almost entirely one-sided. Of the 218
rule-misapplications, 99.5\% (95\% CI [97.4, 99.9]) run in the passive,
risk-averse direction---the stated rule implies \texttt{raise} yet the agent
chooses call, check, or fold---and the violated margin is large rather than
borderline (mean stated equity exceeds the raise threshold by 0.19, and by more
than 0.10 in 72\% of cases). The model computes the inputs and the threshold
correctly but lets hedging language---``marginal strength,'' ``not significantly
above,'' ``a weak hand''---override the quantitative rule it just stated, in a
consistently risk-averse direction. This directional asymmetry, not the share of
any error category, is the robust finding.

\paragraph{An intervention: forbidding hedging halves the override.}
The hedging account makes a testable prediction: instruct the agent to apply the
rule mechanically, and the override should shrink. We added a
\emph{GTO-verbose-decisive} variant---identical to GTO-verbose but with the line
``apply the rule mechanically\ldots do not add qualitative caveats\ldots if
EQUITY > POT\_ODDS + 0.05 you MUST raise, even with a weak-looking hand''---and
ran it against plain GTO-verbose at the same Haiku table (150 hands, two seats
per variant). The instruction roughly halves the failure: the rule-misapplication
rate falls from 13.9\% of decisions to 6.8\%, and overall adherence rises from
60.6\% (95\% CI [57.9, 63.3]) to 68.8\% (95\% CI [66.1, 71.3]). It does not
eliminate it---the residual misapplications remain 96.4\% risk-averse---so the
override is a strong default rather than a mere wording artifact, but it is
partly under prompt control. This both supports the hedging mechanism and offers
a concrete, cheap mitigation.

\paragraph{A second model: the pattern is model-dependent.}
We ran the same diagnostic on DeepSeek-Reasoner, a natively trained reasoning
model, for 200 hands (2{,}036 decisions). Two things differ from Haiku. First,
conclusion$\rightarrow$action consistency is again near-perfect (1.4\% over
1{,}261 tagged decisions). Second, and more important, \emph{rule-misapplication
is rare}: 8.4\% (95\% CI [6.2, 11.2]) of its 466 interpretable errors, versus
33.4\% for Haiku, with the rest split between input errors (51.5\%) and
borderline cases (40.1\%). Equivalently, over the fully parsed decisions in each
run, misapplication accounts for 3.1\% of DeepSeek's against 11.3\% of Haiku's.
The misapplication mechanism that is prominent in Haiku is thus largely absent in
DeepSeek, whose errors instead trace to noisy equity estimates---its stated
equity is within 0.15 of the Monte Carlo value in 64.2\% of decisions, against
68.8\% for Haiku, and four-fifths of its larger misses \emph{under}-state
equity---and to genuinely close spots. The upstream gap is therefore not a single
universal failure mode; its composition depends on the model.\footnote{DeepSeek's
parse rate is lower (62\% vs.\ Haiku's 95\%): even with a raised token budget the
reasoning model sometimes ends its turn before emitting the labeled answer, so
those responses fall back and are excluded. The misapplication estimate is over
parsed responses; given the size of the gap (8\% vs.\ 33\%), plausible exclusion
bias does not close it.}

\subsection{Injecting the Rule Does Not Improve Adherence}

Table~\ref{tab:adherence} reports GTO adherence by prompt strategy across all
three models.
A $\chi^2$ test finds that strategy significantly affects adherence for the two
prompted models (Haiku: $\chi^2 = 8.5$, $p = 0.04$; Flash-Lite: $\chi^2 = 28.9$,
$p < 0.001$) but not for DeepSeek-Reasoner ($\chi^2 = 3.5$, $p = 0.32$). These
tests treat decisions as independent, which overstates significance because
decisions within a hand share board cards, stacks, and opponent states; the
strong Flash-Lite effect is robust to this, but the borderline Haiku result
($p = 0.04$) should be read as suggestive pending hand-level clustering.
For Haiku and Flash-Lite the GTO-constrained strategy never attains the highest
adherence and sits at or near the bottom; for DeepSeek-Reasoner, where the four
strategies do not differ significantly, it happens to rank highest (59.9\%).
In no model, then, does injecting the rule reliably raise adherence above the
alternatives.
Externalizing the decision rule into the prompt therefore does not reliably
improve compliance with that rule. For Haiku this is consistent with the
rule-misapplication mechanism above---the limiting factor is applying a rule, not
knowing it; for DeepSeek-Reasoner, which rarely misapplies the rule, adherence is
instead bounded by noisy equity estimates.
This echoes a recent real-world marketplace experiment in which Claude agents
negotiated on behalf of people \citep{anthropic2025projectdeal}: instructing
agents to negotiate more aggressively produced little change in outcomes, while
weaker agents underperformed in a way that was imperceptible to the people they
represented---a behavioral gap that outcome-level observation did not reveal.

\begin{table}[h]
\centering
\caption{GTO adherence rate by strategy across three model families
(explicit-tag runs, 200 hands each). Cells show adherence \% with decision count
$n$. For Haiku and Gemini the GTO-constrained strategy ranks at or near the
bottom; for DeepSeek-Reasoner the four strategies do not differ significantly
($\chi^2 = 3.5$, $p = 0.32$).}
\label{tab:adherence}
\begin{tabular}{lccc}
\toprule
Strategy & Claude Haiku 4.5 & Gemini 2.5 Flash-Lite & DeepSeek-Reasoner \\
\midrule
Persona          & 55.5\% (265) & 60.4\% (386) & 58.3\% (254) \\
CoT              & 61.0\% (387) & 56.9\% (441) & 54.4\% (419) \\
Baseline         & 53.1\% (343) & 55.9\% (558) & 54.7\% (406) \\
GTO-constrained  & 54.6\% (388) & 49.3\% (499) & 59.9\% (387) \\
\bottomrule
\end{tabular}
\end{table}

\section{Discussion}

\paragraph{Where the gap actually is.}
Decomposing the decision pipeline shows that LLM poker agents are highly
reliable at the conclusion$\rightarrow$action step and noticeably less reliable
at the reasoning$\rightarrow$conclusion step. The upstream weakness is not a
single failure mode, however: errors arise from bad inputs, from misapplying a
correctly stated rule, and from genuinely borderline cases in comparable
measure, and that mix shifts with the model.
For social simulation this is an encouraging and a cautionary message at once:
an agent's final stated decision is a faithful predictor of what it will do, but
the reasoning leading to that decision is not always a valid derivation---and
the dominant way it fails, where it fails at all, is directional (here,
risk-averse) rather than random.
Evaluations that only check whether an action matches a stated conclusion will
see near-perfect fidelity and miss this upstream variability.

\paragraph{Implications for persona modeling and emergent behavior.}
The same decomposition predicts a specific failure for persona-driven agents.
A persona---``tight-aggressive,'' ``risk-averse,'' ``cooperative''---is a
stated disposition meant to shape behavior through reasoning.
Our results show that even a fully stated, correctly computed rationale need
not yield a conclusion consistent with it: in poker, a stated equity rule was
overridden by qualitative hedging in a systematically risk-averse direction.
In a richer social simulation, an analogous override would, we conjecture,
surface as \emph{persona drift}---an agent that articulates its persona
faithfully while its actions are pulled toward a generic, conservative prior
independent of the persona it states.
Our own data offer a small in-domain hint consistent with this: the imposed
tight-aggressive persona barely moves behavior off the no-persona prior, with
Persona adherence (55.5\%) close to Baseline (53.1\%) for Haiku
(Table~\ref{tab:adherence}); imposing a behavioral style changed adherence less
than a percentage point relative to no style at all. This is a single, coarse
observation---adherence is not a direct persona-fidelity measure---but it is the
kind of step-level signal the decomposition is meant to surface.
Whether such drift then produces, in a richer social simulation,
emergent group phenomena---norm formation, coalition dynamics, information
cascades---that reflect a shared prior rather than the configured personas is a
prediction we do not test here; we flag it as a hypothesis, since such drift
would be invisible to outcome-level and conclusion-level checks.
Step-level process-fidelity probes of the kind used here offer one way to
detect such drift before it propagates through a simulation.

\paragraph{Measurement sensitivity.}
The order-of-magnitude difference between the two parsers (Table
\ref{tab:inconsistency}) is itself a finding.
Reported CoT unfaithfulness rates are sensitive to how the conclusion is
elicited and extracted; free-text phrase extraction systematically overstates
inconsistency.
This is a concrete instance of a general concern that results in this area
depend strongly on the elicitation protocol and measurement instrument
\citep{puelmatouzel2026validation}: here, the same behavior measured two ways
differs by an order of magnitude.
We recommend that studies of stated-vs.-actual fidelity elicit an explicit,
machine-checkable conclusion rather than inferring it from prose.

\paragraph{Limitations and future work.}
The three-model scope, while spanning two prompted models and one natively
trained reasoning model (DeepSeek-Reasoner), is still limited.
That the decomposition holds for a native reasoner---whose chain-of-thought is
produced without explicit prompting---is reassuring but not conclusive, and
broader model coverage remains future work.
The rule-misapplication analysis relies on the GTO-verbose elicitation, which
may itself alter behavior relative to silent reasoning. We ran it on two models
(Claude Haiku 4.5 and DeepSeek-Reasoner) and already find the composition of the
upstream gap to be model-dependent; replicating it more broadly is a natural
next step.
Poker is deliberately not a full social simulation: it lacks natural-language
interaction, persuasion, and norm formation, and we do not claim it
substitutes for them.
The contribution is instead methodological. Poker supplies what richer social
simulations lack---a verifiable per-decision reference---and thereby exposes a
step-level fidelity probe (does the stated conclusion follow from the stated
reasoning?) that is otherwise hard to construct.
That probe is portable: any social simulation for which an approximate
reference behavior can be defined can apply the same decomposition.
We therefore treat these experiments as a calibration step toward better
process-fidelity evaluation in social simulation, not as a comprehensive
evaluation of social behavior.

\paragraph{Demo.}
We provide a browser-based demonstration (FastAPI + WebSocket) with three
modes. The \emph{AI Spectator} mode is the interactive counterpart of our
experiment: attendees watch Claude, Gemini, and DeepSeek agents play one
another at a single table, with each agent's per-decision reasoning trace,
explicit stated conclusion, Monte Carlo equity, and executed action surfaced
in real time. This makes both pipeline steps directly observable as they
happen---whether the executed action matches the stated conclusion
(conclusion$\rightarrow$action), and whether that conclusion follows from the
agent's own reasoning (reasoning$\rightarrow$conclusion). Two human-facing
modes reuse the same harness against LLM opponents: \emph{Training} mode shows
the player their equity, pot odds, and a suggested action before each decision;
\emph{Practice} mode delivers a post-hand breakdown of every decision with the
expected-value cost of any deviation from the reference action.
To run it, set an LLM provider API key and launch the bundled web server from
the repository root.
\ifanon
\else
The system and a live demo are available at
\url{https://github.com/louiswang524/texas_poker}.
\fi

\section{Conclusion}

We split the faithfulness gap in LLM poker agents into two steps and found that
they behave very differently.
Conclusion$\rightarrow$action inconsistency is 0.7\% for Haiku and 1.4\% for
DeepSeek once the conclusion is elicited explicitly; the 22--26\% rates reported
by free-text phrase-extraction parsing are a measurement artifact.
The reasoning$\rightarrow$conclusion step is where fidelity frays, but not
through a single dominant failure: the agent's errors split roughly evenly
across bad inputs, rule misapplication, and borderline cases, and this
composition is model-dependent (rule misapplication is a third of Haiku's
interpretable errors but only 8\% of DeepSeek's). The one robust signal is
directional---when an agent misapplies its own restated rule, it almost always
(99.5\% for Haiku) errs risk-aversely---and it is partly hedging behavior:
telling the agent to apply the rule mechanically halves the misapplication rate.
Because these are internal-consistency checks, they are independent of the
reference policy; broader multi-model replication remains future work.
Process-fidelity evaluation for LLM-based social simulation should therefore
target the derivation of conclusions from reasoning, and should elicit
machine-checkable conclusions to avoid conflating measurement noise with model
behavior.

\bibliography{colm2026_conference}

@inproceedings{wei2022chain,
  title     = {Chain-of-thought prompting elicits reasoning in large language models},
  author    = {Wei, Jason and Wang, Xuezhi and Schuurmans, Dale and Bosma, Maarten and Chi, Ed and Le, Quoc and Zhou, Denny},
  booktitle = {Advances in Neural Information Processing Systems},
  volume    = {35},
  year      = {2022}
}

@inproceedings{turpin2023language,
  title     = {Language models don't always say what they think: Unfaithful explanations in chain-of-thought prompting},
  author    = {Turpin, Miles and Michael, Julian and Perez, Ethan and Bowman, Samuel R.},
  booktitle = {Advances in Neural Information Processing Systems},
  volume    = {36},
  year      = {2023}
}

@article{lanham2023measuring,
  title   = {Measuring faithfulness in chain-of-thought reasoning},
  author  = {Lanham, Tamera and Chen, Anna and Radhakrishnan, Ansh and Steiner, Benoit and Denison, Carson and Hernandez, Danny and Li, Dustin and Durmus, Esin and Hubinger, Evan and Kernion, Jackson and others},
  journal = {arXiv preprint arXiv:2307.13702},
  year    = {2023}
}

@inproceedings{park2023generative,
  title     = {Generative agents: Interactive simulacra of human behavior},
  author    = {Park, Joon Sung and O'Brien, Joseph C. and Cai, Carrie J. and Morris, Meredith Ringel and Liang, Percy and Bernstein, Michael S.},
  booktitle = {Proceedings of the ACM Symposium on User Interface Software and Technology (UIST)},
  year      = {2023}
}

@inproceedings{zhou2024sotopia,
  title     = {{SOTOPIA}: Interactive evaluation for social intelligence in language agents},
  author    = {Zhou, Xuhui and Zhu, Hao and Mathur, Leena and Zhang, Ruohong and Yu, Haofei and Qi, Zhengyang and Morency, Louis-Philippe and Bisk, Yonatan and Fried, Daniel and Neubig, Graham and Sap, Maarten},
  booktitle = {International Conference on Learning Representations (ICLR)},
  year      = {2024}
}

@article{costarelli2024gamebench,
  title   = {{GameBench}: Evaluating strategic reasoning abilities of {LLM} agents},
  author  = {Costarelli, Anthony and Allen, Mat and Hauksson, Roman and Sodunke, Grace and Hariharan, Suhas and Cheng, Carlson and Li, Wenjie and Clymer, Joshua and Yadav, Arjun},
  journal = {arXiv preprint arXiv:2406.06613},
  year    = {2024}
}

@inproceedings{lin2026far,
  title     = {How far are {LLMs} from professional poker players? {R}evisiting game-theoretic reasoning with agentic tool use},
  author    = {Lin, Minhua and Dai, Enyan and Liu, Hui and Tang, Xianfeng and Yan, Yuliang and Dai, Zhenwei and Zeng, Jingying and Zhang, Zhiwei and Wang, Fali and Gao, Hongcheng and Luo, Chen and Zhang, Xiang and He, Qi and Wang, Suhang},
  booktitle = {International Conference on Learning Representations (ICLR)},
  year      = {2026}
}

@article{xie2026m3bench,
  title   = {{M3-BENCH}: Process-aware evaluation of {LLM} agents' social behaviors in mixed-motive games},
  author  = {Xie, Sixiong and Shi, Zhuofan and Shen, Haiyang and Ma, Yun and Jing, Xiang},
  journal = {arXiv preprint arXiv:2601.08462},
  year    = {2026}
}

@inproceedings{puelmatouzel2026validation,
  title     = {Position: Time to Close the Validation Gap in {LLM} Social Simulations},
  author    = {Puelma Touzel, Maximilian and Sarangi, Sneheel and B\"uck-Kaeffer, Aur\'elien and Yang, Zachary and Godbout, Jean-Fran\c{c}ois and Rabbany, Reihaneh},
  booktitle = {Proceedings of the 43rd International Conference on Machine Learning (ICML)},
  series    = {PMLR 306},
  year      = {2026}
}

@misc{anthropic2025projectdeal,
  title        = {Project Deal},
  author       = {{Anthropic}},
  year         = {2025},
  howpublished = {\url{https://www.anthropic.com/features/project-deal}}
}
\bibliographystyle{colm2026_conference}

\appendix

\section{Prompts}
\label{app:prompts}

Table~\ref{tab:prompts} lists the verbatim system prompts for all five
strategies (the shared expert framing is elided with ``\ldots'') together with
an example of the per-decision user prompt. Only the system prompt varies
across strategies; every strategy receives the same game-state user prompt.

\begin{table}[h]
\centering
\footnotesize
\begin{tabular}{@{}p{0.96\columnwidth}@{}}
\toprule
\textbf{Baseline} \\
\ttfamily You are an expert Texas Hold'em poker player. Your goal is to
accumulate chips by making +EV decisions. You must respond with valid JSON
only --- no reasoning, no explanation, no other text. Example:
\{"action": "call", "amount": 0\} \\
\midrule
\textbf{CoT} \\
\ttfamily \ldots Think step by step: assess your hand strength, estimate pot
odds, consider opponents. After your reasoning, state your conclusion on its
own line in EXACTLY this format: DECISION: <fold|check|call|raise>. Then on the
very last line respond with JSON only. \\
\midrule
\textbf{Persona} \\
\ttfamily You are a tight-aggressive (TAG) Texas Hold'em player\ldots You play
only premium hands aggressively, fold marginal hands, and apply pressure from
position. You must respond with valid JSON only\ldots \\
\midrule
\textbf{GTO-constrained} \\
\ttfamily \ldots Mentally estimate equity vs pot odds (pot\_odds = call\_amount
/ (pot + call\_amount)). Fold if equity < pot\_odds; call or raise if equity >
pot\_odds. You must respond with valid JSON only\ldots \\
\midrule
\textbf{GTO-verbose} (diagnostic) \\
\ttfamily \ldots Show your work in EXACTLY these labeled lines, then the JSON:
EQUITY: <0.0--1.0>; POT\_ODDS: <0.0--1.0>; RULE: fold if EQUITY < POT\_ODDS;
raise if EQUITY > POT\_ODDS + 0.05; otherwise call; DECISION:
<fold|check|call|raise> \\
\midrule
\textbf{GTO-verbose-decisive} (ablation) \\
\ttfamily \ldots Apply the rule MECHANICALLY to your own numbers. Do NOT add
qualitative caveats about hand quality, position, or playability, and do NOT
second-guess the rule: if EQUITY > POT\_ODDS + 0.05 you MUST raise, even with a
weak-looking hand. [then the same labeled lines as GTO-verbose] \\
\midrule
\textbf{Example user prompt (shared by all strategies)} \\
\ttfamily Street: PREFLOP. Your hole cards: A$\spadesuit$ K$\spadesuit$.
Community cards: none. Pot: 15. Current bet to call: 10. Your stack: 9990.
Players: AI-1 stack=9990 active; AI-2 stack=10000 folded; AI-3 stack=9995
active. Available actions: call (10 chips), fold, raise (min 20 chips). \\
\bottomrule
\end{tabular}
\caption{System prompts for the five strategies and an example per-decision
user prompt.}
\label{tab:prompts}
\end{table}

\end{document}